**Wireless bioelectronic control architectures for biohybrid robotic systems**

*Hiroyuki Tetsuka* and Minoru Hirano*

Frontier Management Office, Toyota Central R&D Labs., Inc., 41-1 Yokomichi, Nagakute, Aichi 480-1192, Japan

E-mail: h-tetsuka@mosk.tytlabs.co.jp

**Abstract.**

Wireless bioelectronic interfaces are increasingly used to control tissue-engineered biohybrid robotic systems. However, a unifying engineering framework linking device design to system-level control remains underdeveloped. Here, we propose that wireless control in biohybrid robotics can be formulated as a coupled co-design problem of integrating signal delivery, spatial selectivity, scalability, and interface stability. We analyze three representative control strategies, wireless electrical stimulation, wireless optoelectronic stimulation, and neuromuscular integration, which operates within a distinct regime with characteristic trade-offs. Across these modalities, the tissue–device interface emerges as a key constraint, governing the interplay between electromagnetic coupling, circuit performance, and biomechanical response. Based on this framework, we outline practical design principles spanning electromagnetic field distribution, circuit architecture, and actuator mechanics. We further propose a transition from open-loop stimulation to closed-loop biohybrid autonomy enabled by organoid-integrated bioelectronics and bidirectional microelectrode interfaces. This work establishes a system-level perspective on wireless bioelectronic control and provides design guidelines for developing stable, scalable, and autonomous biohybrid robotic systems.

## 1. Introduction

Biohybrid robots embed living muscles—typically cardiac or skeletal muscle tissues—into highly pliable structures to create living machines capable of soft, adaptive behaviors like a living organism. These robots can mimic biological behaviors and functions such as a life-like stimulus–response dynamics; however, achieving reliable control remains challenging. Conventional electrical pacing generally relies on tethered wiring or bulky electrodes immersed in culture media, which limits applicability in enclosed or constrained environments, such as in vivo environments.

Over the past two decades, the field has advanced from early proof-of-concept demonstrations of living muscle-powered microdevices and thin-film actuators to increasingly sophisticated biomimetic robots capable of swimming, walking, and multifunctional actuation (1-19). Fig. 1 reframes it as an evolution of control strategies that converges on closed-loop autonomy. Early studies emphasized muscle alignment to improve anisotropic force generation and structural control. Subsequent work introduced optogenetic control to enable programmable remote activation of muscle tissues. More recent systems incorporated wireless bioelectronic and neuromuscular control, improving compatibility with enclosed environments and enabling selective actuation through multiplexed stimulation and neuron-mediated pathways. In this perspective, these stages are interpreted as successive layers of control complexity that motivate the future integration of bidirectional sensing, adaptive control, and biological information processing. Recent advances can be organized into three major control directions: (i) wireless bioelectronics, (ii) wireless optoelectronics, and (iii) neuromuscular integration, each of which reflects broader trends in biohybrid robotics. We then distill practical co-design principles across these strategies to outline a future route toward autonomous systems enabled by neural organoid-wireless bioelectronic integration.

To contextualize the evolution in biohybrid robotics, key milestones can be interpreted as successive advances in controllability and system integration. The earliest muscle-powered microdevices demonstrated that contractile tissues can integrate with microdevices and function at the microscale (1). Subsequent muscle alignment control strategies established biomimetic bioactuation architectures suitable for engineered locomotion (2, 3). Optogenetic, fish-inspired systems then introduced spatially programmable muscle activation through the programmable light illumination, enabling gait control of robots (4, 6). Microelectrode-integrated ray platforms further highlighted the importance of embedded electrical stimulation interfaces in controlling locomotion (5). The integration of wireless bioelectronics then replaced tethered wiring and increased compatibility with closed environments (7). Neuromuscular bioelectronic robots further enabled selective actuation through multiplexed wireless control and neuron-mediated transduction, allowing steering of robot (8, 9, 10, 11). Skeletal-muscle-based robots progressed along a distinct yet complementary route. The first skeletal muscle-powered bio-bots demonstrated electrically paced, walking-like locomotion (12). Subsequent optogenetic bio-bots introduced optical programmability (13), while neuromuscular robots mimicked neural innervation of biological muscles (15). Wireless optoelectronic robots later achieved battery-free, remote control of multifunctional tasks (16). An optoelectronic neuromuscular robot further combined micro-light-emitting diode (uLED)-based wireless optoelectronics with neuromuscular pathways, illustrating a strategy for selective, multi-site activation (19).

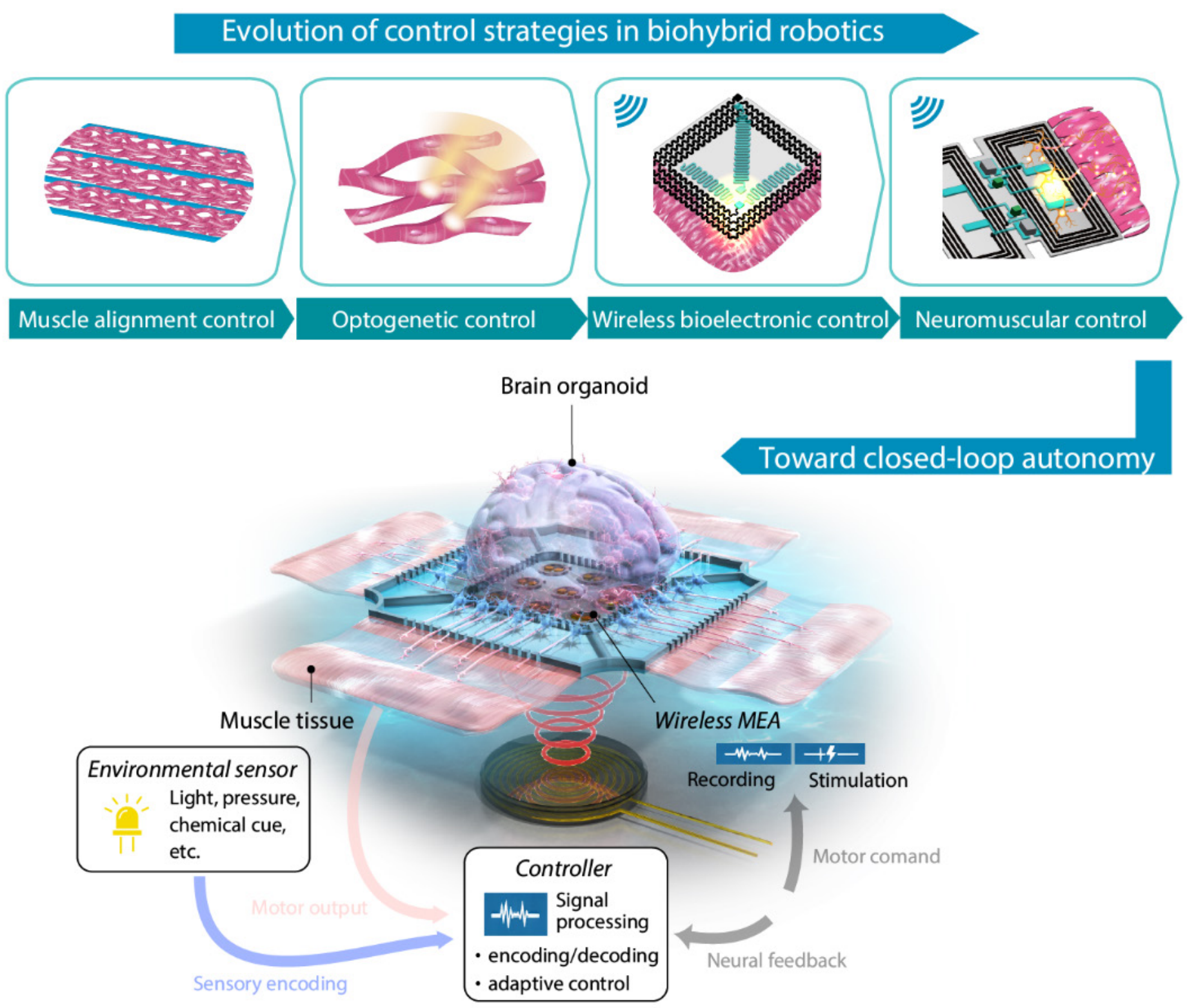

**Figure 1.** Evolution of control strategies in biohybrid robotics toward closed-loop autonomy. Top: representative progression from muscle alignment control to optogenetic control, wireless bioelectronic control, and neuromuscular control. Bottom: forward-looking concept of a closed-loop autonomous biohybrid robot in which environmental signals are encoded by a controller, delivered through a wireless microelectrode array (MEA) to a brain organoid, and decoded into motor commands that regulate a muscle tissue-based body.

## 2. Wireless control strategies

Table 1 summarizes the major directions, namely, wireless bioelectronics (wireless electrical stimulation), wireless optoelectronics (wireless optical stimulation), neuromuscular integration, and a future co-integration concept of neural organoid–wireless bioelectronics. Wireless bioelectronics offers the most direct and efficient hardware route to untethered muscle stimulation. However, its selectivity is often limited because stimulation is typically governed by global fields. On the other hand, wireless optoelectronics improves spatial addressability through μLED placement and patterned illumination. But it still has several practical challenges, including the need for optogenetic modification, loss of light intensity in tissue, and heat generation. Neuromuscular integration further expands controllability by exploiting biological signal transduction, enabling selective multi-actuator control through multiplexed neural stimulation. However, their performance becomes more dependent on tissue maturation, synapse modality, and system-level complexity. In this sense, progression across these directions can be understood as a trade-off from simplicity and robust pacing toward higher selectivity, richer control, and ultimately compatibility with closed-loop autonomy.

The future concept of neural organoid–MEA integration should be viewed as a forward-looking extension of current organoid electrophysiology and biocomputing studies, rather than as a capability already established in current technologies. Relevant precedents include closed-loop learning and control in in vitro neuronal systems interfaced through MEAs (20), high-density organoid electrophysiology (21), cerebral organoid and assembloid interfaces enabled by kirigami electronics (22), and organoid-intelligence frameworks for biocomputing (23). In addition, interface platforms that provide broader access to organoid activity, including 360° size-adjustable MEA architectures with multichannel coverage, may offer a more suitable basis for organoid-based information processing than limited-contact planar configurations (24). Nevertheless, most existing demonstrations remain wired, and achieving stable wireless, high-density, stable, bidirectional organoid–MEA interfacing remains a major engineering challenge.

**Table 1.** Comparative overview of wireless control strategies for untethered biohybrid robots and a future direction toward neural organoid MEA autonomy.

| Direction | Signal path | Selectivity / addressability | Scale-up path | I/O | Closed-loop capability | Typical stability | Bottlenecks / | Refs. |
|---|---|---|---|---|---|---|---|---|
| (1) Wireless bioelectronics (electrical stimulation) | RF field -> receiver -> rectification/pulsing -> direct electrical pacing | Low-moderate; usually 1 global field or 1 local pacing site per receiver; field uniformity sets selectivity | Usually 1 receiver:1 output; scaling limited by coil coverage, standoff distance, and field nonuniformity | Usually stimulation-only; sensing/telemetry added separately | Open-loop in representative systems; closed-loop needs added sensing + control | ~days-weeks in aqueous media | Media-appropriate electromagnetic and circuit co-design is essential; avoid overstimulation and electrochemical limits | (7) |
| (2) Wireless optoelectronics (wireless optical stimulation) | Wireless power -> micro-LED/uLED emission -> optogenetic activation | Moderate-high; 1-several emitter sites defined by emitter placement | Add emitters/addressing; limited by power budget, heat dissipation, and encapsulation | Mainly stimulation; communication/telemetry possible | Programmable open-loop; closed-loop only when telemetry is added | Acute to short-term / multi-day; longer use limited by heat, encapsulation, and opsin stability | Optical loss, wavelength, irradiance duty cycle, emitter topology is a primary lever for selectivity | (4) (13) (16) (19) |
| (3) Neuromuscular integration | Wireless power/commands -> selective neural stimulation -> biological | High; 2 actuator arrays independently addressed at 6.78 and 13.56 MHz | 2 channels demonstrated; more via FDM/TDM, with higher spectral/control complexity | Mainly stimulation; recording can be added; biological transduction at the NMJ | High potential; steering/speed control shown, but integrated sensing is still limited | >150 s sustained pacing in one representative system; multi-week functional | Synapse modality and tissue maturation matter; multiplexing strategy shapes gait and steering | (8) (9) (10) (11) (15) (19) |

| | transduction to muscle | | | | | retention reported | | |
|---|---|---|---|---|---|---|---|---|
| (4) Future: Organoid + wireless MEA | Bidirectional MEA: stimulation input + recording output, ideally wireless in sealed systems | Very high; multi-site stimulation + recording; MEA systems can span tens-thousands of electrodes/channels | Limited by channel count, data bandwidth, decoding complexity, and chronic interface stability | Bidirectional by design: multi-channel recording + multi-site stimulation | Intrinsic fit for closed-loop control; in vitro closed-loop learning/control shown, but wireless organoid-MEA remains prospective | Weeks-months in wired/chronic organoid-MEA interfaces; long-term wireless stability not yet shown | Synapse modality and tissue maturation matter, plasticity drift | |

As shown in Fig. 2, these directions can be expressed within a common systems-level architecture, where the key differentiators are (i) where the stimulus is generated, electrodes versus light emitters, (ii) whether biological transduction is leveraged, muscle-only versus neuromuscular pathways, and (iii) whether bidirectional I/O is available to support closed-loop control, such as microelectrode array (MEA)-based recording and stimulation.

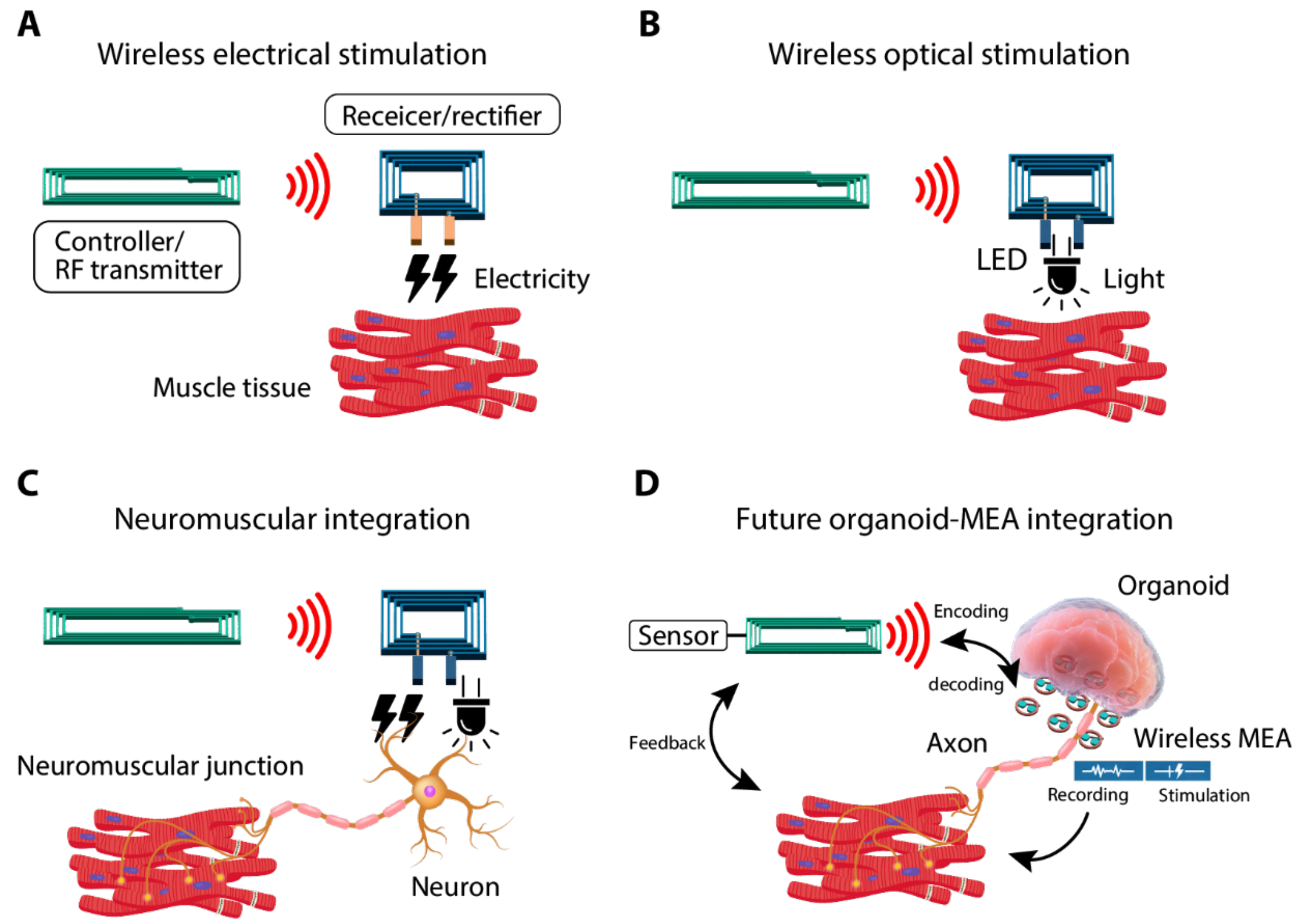

**Figure 2.** System-level architectures of wireless biohybrid robot control across three streams and a future organoid and MEA direction. (A) Wireless electrical stimulation: inductive radio frequency power transfer drives a receiver and rectifier that generate pulsed electrical output delivered through tissue-facing electrodes to directly pace muscle tissues. (B) Wireless optoelectronics: wireless power drives micro light-emitting diode (μLED) emitters to deliver light for optogenetic activation of muscle or neurons, enabling spatial selectivity through emitter placement. (C) Neuromuscular integration with selective wireless control: wireless addressing, such as multiplexed excitation, targets neural tissues coupled to muscle via neuromuscular or neurocardiac junctions, leveraging biological transduction to shape actuation and enabling independent control of multiple actuators. (D) Future direction: organoid and wireless MEA autonomy. A bidirectional interface will record organoid activity and deliver stimulation patterns. This will enable the decoding and encoding of pathways for closed-loop control of muscle-based actuation in closed environments.

### 2.1 Wireless bioelectronics

Wireless electrical stimulation is a core approach in wireless bioelectronics because it preserves the direct control of electrical pacing without relying on wired connections, and it remains one of the most established control routes in tissue-based biohybrid robotics (26, 27). Its advantage is hardware simplicity and reliable pacing in closed and aqueous environments. Main limitations are signal selectivity, electric field nonuniformity, and electrochemical safety constraints at the tissue–electrode interface. One platform has shown that wirelessly powered, stretchable, lightweight bioelectronics can be embedded directly into a hydrogel-based muscle tissue construct. The system delivers controlled monophasic pulses up to ~9 V and has been used to remotely trigger locomotion in robots (7). A recent study further suggests that wireless bioelectronic devices can be miniaturized within a footprint of ~23 $mm^2$ while preserving effective electrical stimulation. The system generated distance-dependent output voltage pulses of ~2–6 V, enabling wireless pacing of robot. This miniaturization serves as a practical route toward compact and media-compatible closed-system biohybrid robotics (25).

A central design consideration is the co-design of the electric field distribution, circuit, and muscle tissue alignment in aqueous environments. The output voltage in culture media depends strongly on the surrounding dielectric environment, electrode impedance, and transmitter–receiver distance. The performance of wireless bioelectronic platform drops rapidly with standoff distance (7). Electrical stimulation in conductive media is also sensitive to field nonuniformity and local current density. Electrode size, spacing, and placement should therefore be selected not only to exceed the stimulation threshold but also to avoid localized hotspots, overstimulation, or unstable pacing. This matters because the usable operating range is limited by stimulation thresholds, electrochemical safety constraints, and the mechanical bandwidth of the scaffold–tissue actuator, all of which have been identified as constraints in biohybrid robot design (26, 27).

Validation should be done in the intended operating environment. Radio-frequency coupling efficiency and the resulting output waveforms can differ markedly between air and liquid. Coil geometry and drive conditions therefore need to be optimized together, with explicit safety margins, so that stimulation thresholds are met across the workspace while minimizing localized hotspots that could cause overstimulation. Even when adequate electrical pacing is achievable, controllability at the robot level is often limited by muscle relaxation dynamics and the viscoelastic response of hydrogel or scaffold composites. As a result, mechanical bandwidth sets the practical upper limit on actuation frequency and should be treated as a primary design constraint.

### 2.2 Wireless optoelectronics

Wireless optical stimulation can be achieved using wireless optoelectronics by integrating light sources that activate optogenetically modified muscle tissue. Wireless optoelectronics extends remote control beyond direct electrical pacing by introducing spatially programmable optical stimulation. Its major strength is improved addressability, because stimulation sites can be defined by μLED placement and illumination patterns rather than by electric-field distribution alone. At the same time, this advantage introduces additional constraints, including the need for optogenetic modification, optical attenuation in tissue and media, thermal management, encapsulation reliability, and long-term opsin stability (28, 29). Previous studies have reported that optoelectronic robots incorporating battery-free μLEDs support wireless control and real-time communication (16). Optical stimulation offers spatial selectivity through emitter placement and decouples actuation from direct electrode-tissue contact.

From a design perspective, effective optoelectronic control requires co-design between the biophysical properties of light-sensitive opsins and device hardware, including appropriate choices of wavelength, irradiance, and illumination strategy for the target tissue (28). Light wavelength, irradiance intensity, and duty cycle directly affect activation thresholds, temporal fidelity, and optical penetration depth. In practical operation, optical attenuation in the culture medium and the tissue itself changes the delivered light dose. Thus, the effective intensity at the actuator may differ significantly. Thermal management is also essential, because sustained illumination can produce biologically relevant heating that perturbs tissue function and limits long-term operational stability (29). In addition, potential phototoxic and photochemical effects associated with prolonged high-intensity illumination should be considered when designing long-term stable systems (28, 29).

### 2.3 Neural integration

An integration of nervous system can move beyond biologically mediated control to enable selective control of multiple actuators and mimic biological signal pathway. The major strength of this approach is enhanced selectivity through multiplexed neural pathways. However, the main challenges lie in tissue maturation, synaptic reliability, and the increased complexity of maintaining stable neuromuscular function. One notable approach uses wireless frequency multiplexing to selectively stimulate neural tissue connected to muscle actuators. In the representative system, two modulation signals were induced in separate resonant coils to generate voltage pulses that stimulate the neural tissues. This configuration enabled the independent modulation of two actuator arrays, providing control over the robots' speed and steering (8). This work also demonstrates the robustness of their signal transduction. Controlled pacing mediated through electrical synapses was sustained longer than chemical synapses.

More recently, neuromuscular junction–based biohybrid crawling robots integrated optogenetically-modified motor neurons and skeletal muscle with onboard wireless μLED optoelectronics, enabling optical stimulation of neural tissue to modulate neuromuscular actuation (19). By shifting the site of stimulation from muscle to neurons, optoelectronic control expands the available actuation repertoire and reduces reliance on direct muscle pacing. This may be beneficial when repeated direct stimulation of muscle is undesirable because it may affect muscle integrity and limit long-term operation.

## 3. Co-design principles

The design of remote-controlled biohybrid robots should be guided by coupled factors such as signal/power transfer, selectivity, and mechanics across the control strategies described above. A central trade-off lies between signal/power transfer and selectivity. Wireless electrical pacing prioritizes efficient transfer of signals and power to tissue-electrode interfaces. Wireless optoelectronics distribute delivered power across spatially arranged emitters to achieve addressability. Neuromuscular integration enhances selectivity further by exploiting biological transduction and timing control, particularly when combined with multiplexing schemes. In parallel, packaging and interface stability in aqueous environments often dominate system performance. Encapsulation and electrode performance integrity become especially limiting under motion, and electrical excitation conditions must also be tuned to avoid localized overstimulation. Finally, actuation and locomotion remain constrained by tissue and structure, as scaffold–tissue composites often limit achievable amplitude and actuation frequency through relaxation dynamics and viscoelastic properties. As such, system performance emerges from the coupled design of fields and circuits, interfaces in media, and the mechanical properties of scaffolds.

To control actuator arrays, multiple muscle actuators or multiple regions of a single actuator must be independently activated. This requires scalable addressing schemes such as frequency multiplexing, time-division strategies, and/or spatial optical addressing. As illustrated in Fig. 3, control of scalable actuator arrays relies on addressing approaches that trade off selectivity, power efficiency, and hardware complexity. Frequency multiplexing provides selectivity by matching excitation bands to multiple resonant receivers, whereas time-division multiplexing and coded stimulation increase channel count without additional physical interconnects. Spatial addressing, implemented for optically through μLED placement or electromagnetically through field shaping, offers an alternative path to selectivity but often introduces added constraints on packaging and integration.

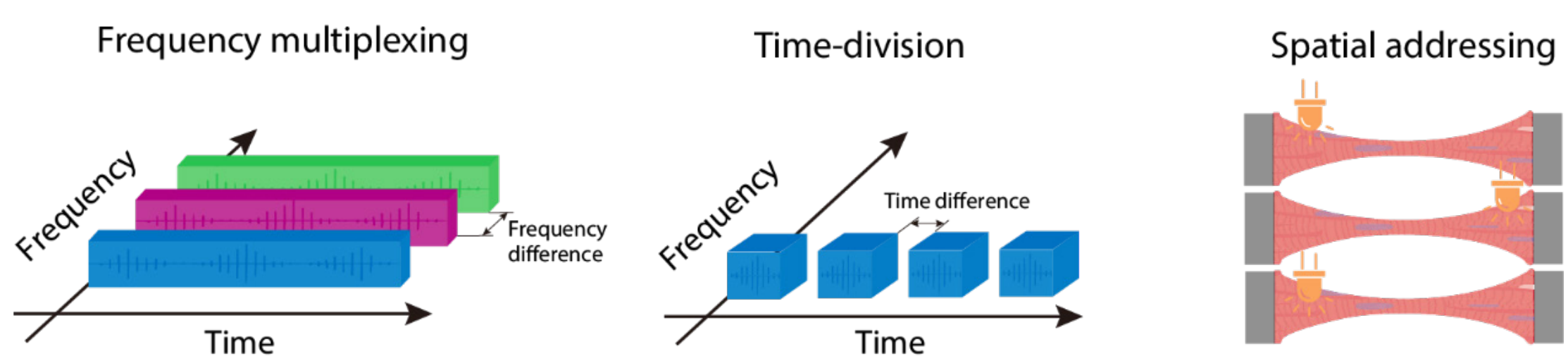

**Figure 3.** Control strategies for scalable multiple actuator arrays. Frequency multiplexing in which resonant receivers with different center frequencies enable selective drive of multiple stimulation electrodes. Time-division multiplexing in which temporal scheduling or programmed waveforms offer addressability within a shared carrier band. Spatial addressing in which selectivity activation through emitter placement, such as μLED arrays, or through electrode arrays.

## 4. Challenges and opportunities

These co-design constraints give rise to several challenges and opportunities. Miniaturization reduces the coupling area and consequently the energy delivered at the interface. This motivates the development of higher-efficiency resonant receivers, improved electromagnetic field uniformity during excitation, and interface strategies that effectively lower stimulation thresholds. One way to achieve this may be to move stimulation upstream within the neuromuscular system. For long-term use, the electronics must remain reliable while tissue health is maintained, making chronic packaging integrity, biological integration, and long-term biocompatibility central design requirements (27, 30). In addition, most current platforms operate in open loop with preset stimulation schedules. Reaching autonomous operation will therefore require integrated sensing and feedback, along with adaptive encoding and decoding that take advantage of biological dynamics, consistent with the broader architecture of electronic neural interfaces that combine tissue interfacing, sensing, and signal processing (31). Together, these capabilities enable closed-loop control.

As illustrated in the lower panel of Fig. 1, the historical evolution summarized in the upper row points to a roadmap from early muscle-powered prototypes toward neural organoid-enabled autonomous systems with multiple muscle arrays. A compelling next step is the emergence of autonomous biohybrid robots in which a biological neural construct functions as an information-processing core that interacts bidirectionally via wireless bioelectronics. By incorporating the central nervous system, autonomy arises from closed-loop interaction rather than preprogrammed open-loop stimulation. In this framework, wireless bioelectronics serve as bidirectional interfaces, linking stimulation, recording, and downstream signal processing within a unified neural-interfacing architecture (31). Environmental cues are encoded into biological neural activity and decoded into motor commands. A brain organoid or neural network provides adaptive dynamics and plasticity, while wireless bioelectronics, such as a wireless MEA, establish programmable input/output pathways. Stimulation patterns act as sensory surrogates, and recorded neural activity is decoded to generate motor commands. Conceptually, the robot becomes a closed loop.

This direction is motivated by several converging advances: (i) demonstrations of closed-loop learning and control using in vitro neuronal systems interfaced via MEAs; (ii) high-density, MEA-based electrophysiological recordings of organoids sustained over weeks; (iii) chronic recordings from organoids and assembloids enabled by kirigami electronics over months; and (iv) organoid intelligence frameworks for biocomputing. In neural organoid–wireless MEA autonomy, bidirectional input/output (I/O) is the core requirement. However, the realization of fully wireless MEA systems capable of stable, high-density, and chronic bidirectional interfacing remains a significant engineering challenge, particularly with respect to power delivery, data bandwidth, encapsulation, and long-term biocompatibility (30, 31). Moreover, current demonstrations of organoid–MEA integration largely rely on wired platforms, and translation to wireless architectures will require careful co-optimization of electronics, packaging, and tissue stability. Beyond hardware considerations, meaningful closed-loop autonomy cannot be assumed to arise from simple physical integration of neural organoids with muscle actuators. Spontaneous organoid activity does not inherently guarantee structured or task-specific motor output. Achieving functional control will therefore depend critically on engineered encoding–decoding strategies and adaptive control frameworks. These interpret neural population dynamics rather than relying on intrinsic organization alone, especially if organoids are to be used as computational substrates rather than merely as spontaneously active tissues (32). These considerations highlight the need to move beyond simple integration and toward clearly defined, measurable system-level performance criteria. Therefore, chronic recording quality, stimulation safety, and long-term stability should be considered primary performance metrics. Early-stage demonstrations will likely benefit from low-dimensional control, such as regulating rate and locomotion speed according to environment. Designs must explicitly account for nonstationarity: organoid maturation, plasticity, and drift will shift baselines over time, so controllers and encoding/decoding strategies should be adaptive rather than assuming fixed neural dynamics, particularly in organoid-based computing systems in which stimulation-dependent reshaping of network dynamics is expected (32).

## 5. Conclusion

Biohybrid robotics has progressed through identifiable phases of increasing controllability, from cellular alignment strategies to optogenetic control and, more recently, to wireless bioelectronic and neuromuscular control phases, as summarized in Fig. 1. Wireless electrical stimulation enables robust untethered pacing, delivering monophasic pulses in aqueous environments (7). Wireless optoelectronics adds optical control channels via onboard wireless uLED systems have enabled optogenetic neuromuscular activation (16, 19). Neuromuscular integration combined with frequency multiplexing supports selective multi-actuator control and maintains robust signal pathways over extended durations (8). Across these approaches, the central trade-off is between simplicity and controllability: direct wireless electrical pacing is the most straightforward to implement, optoelectronics improves spatial selectivity, and neuromuscular integration offers the richest route to selective and adaptive actuation with greater biological and system complexity.

These progressions motivate the future transition toward bidirectional organoid–MEA systems for closed-loop autonomy. At present, however, most biohybrid robotic systems remain experimental in vitro platforms, operating in culture environments with limited force output, scalability, and environmental robustness. As such, they are not yet suited for stand-alone deployment or long-range autonomous exploration. Their near-term impact is more likely to lie in serving as controlled experimental testbeds for studying neuromuscular integration, embodied control, and bioelectronic interfacing.

Looking forward, the realization of stable wireless MEA platforms capable of long-term bidirectional interfacing remains a key engineering challenge. If achieved, integrating wireless MEA platforms with brain organoids offers a pathway to closed-loop autonomy, where biological computation and engineered bioelectronics jointly shape behavior. Together, these converging streams outline a roadmap from untethered pacing, to selective actuator addressing, and ultimately to autonomous biohybrid robots capable of adapting to their environment through embodied neural feedback.